\documentclass[conference]{IEEEtran}
\IEEEoverridecommandlockouts
\usepackage{tabularx}
\usepackage{cite}
\usepackage{amsmath,amssymb,amsfonts}
\usepackage{algorithmic}
\usepackage{graphicx}
\usepackage{textcomp}
\usepackage{xcolor}
\usepackage{makecell}
\usepackage{enumitem}

\makeatletter
\renewcommand{\IEEEauthorrefmark}[1]{\textsuperscript{#1}}
\makeatother

\def\BibTeX{{\rm B\kern-.05em{\sc i\kern-.025em b}\kern-.08em
    T\kern-.1667em\lower.7ex\hbox{E}\kern-.125emX}}
\begin{document}

\title{Align then Refine: Text-Guided 3D Prostate Lesion Segmentation\\
}

\author{
\IEEEauthorblockN{
Cuiling Sun\IEEEauthorrefmark{1},
Linkai Peng\IEEEauthorrefmark{1},
Adam Murphy\IEEEauthorrefmark{1},
Elif Keles\IEEEauthorrefmark{1},
Hiten D. Patel\IEEEauthorrefmark{2},
Ashley Ross\IEEEauthorrefmark{2},
Frank Miller\IEEEauthorrefmark{1},\\
Baris Turkbey\IEEEauthorrefmark{3},
Andrea Mia Bejar\IEEEauthorrefmark{1},
Halil Ertugrul Aktas\IEEEauthorrefmark{1},
Gorkem Durak\IEEEauthorrefmark{1},
Ulas Bagci\IEEEauthorrefmark{1}
}
% \\[0.2ex]
\IEEEauthorblockA{\IEEEauthorrefmark{1}Department of Radiology, Northwestern University, Chicago, USA}
\IEEEauthorblockA{\IEEEauthorrefmark{2}Department of Urology, Northwestern University, Chicago, USA}
\IEEEauthorblockA{\IEEEauthorrefmark{3}Center for Cancer Research, National Cancer Institute, Bethesda, USA}
% \\[0.5ex]
\IEEEauthorblockA{\textit{Emails:} CuilingSun2026@u.northwestern.edu, ulas.bagci@northwestern.edu}
}

\maketitle

\begin{abstract}
Automated 3D segmentation of prostate lesions from biparametric MRI (bp-MRI) is essential for reliable algorithmic analysis, but achieving high precision remains challenging. Volumetric methods must combine multiple modalities while ensuring anatomical consistency, but current models struggle to integrate cross-modal information reliably. While vision-language models (VLMs) are replacing the currently used architectural designs, they still lack the fine-grained, lesion-level semantics required for effective localized guidance. To address these limitations, we propose a new multi-encoder U-Net architecture incorporating three key innovations: (1) an alignment loss that enhances foreground text-image similarity to inject lesion semantics; (2) a heatmap loss that calibrates the similarity map and suppresses spurious background activations; and (3) a final-stage, confidence-gated multi-head cross-attention refiner that performs localized boundary edits in high-confidence regions. A phase-scheduled training regime stabilizes the optimization of these components. Our method consistently outperforms prior approaches, establishing a new state-of-the-art on the PI-CAI dataset through enhanced multimodal fusion and localized text guidance. Our code is available at https://github.com/NUBagciLab/Prostate-Lesion-Segmentation.
\end{abstract}


\section{Introduction}
% Accurate volumetric segmentation of prostate lesions from biparametric MRI (bp‑MRI)—comprising T2‑weighted (T2W), diffusion‑weighted imaging (DWI), and apparent diffusion coefficient (ADC) sequences—is central to robust algorithmic analysis and downstream clinical assessment. Yet the task remains challenging in practice: lesions are frequently small, heterogeneous, and poorly contrasted against surrounding tissue; modality‑specific artifacts, misregistration, and cross‑scanner domain shifts further complicate fusion in three dimensions~\cite{Gunashekar2024}. These factors degrade boundary fidelity and undermine the reliability required for patient‑level decision support.

Accurate volumetric segmentation of prostate cancer lesions from biparametric MRI (bp-MRI)—comprising T2-weighted (T2W), diffusion-weighted imaging (DWI), and apparent diffusion coefficient (ADC) sequences—plays a critical role in prostate cancer diagnosis, risk stratification, and treatment planning. Precise lesion delineation directly impacts targeted biopsy guidance, focal therapy planning, and longitudinal disease monitoring, and errors in segmentation may lead to missed clinically significant tumors or inaccurate assessment of tumor burden.

Despite its importance, reliable prostate lesion segmentation remains challenging in practice. Lesions are frequently small, heterogeneous, and poorly contrasted against surrounding tissue; modality-specific artifacts, misregistration, and cross-scanner domain shifts further complicate three-dimensional fusion~\cite{Gunashekar2024}. These factors degrade boundary fidelity and undermine the robustness required for patient-level clinical decision support.

Existing approaches fall into three broad families. First, \textbf{image‑only} architectures, U‑Net and transformer‑based decoders, provide strong baselines but typically rely on fixed, reliability‑agnostic multimodal fusion and often process 3D volumes as 2D stacks, sacrificing across‑slice context and stability in the presence of misalignment~\cite{nnUNet, 3dunet, hatamizadeh2022swinunetrswintransformers, hatamizadeh2021unetrtransformers3dmedical, oktay2018attentionunetlearninglook}. Second, \textbf{promptable foundation models} adapted to medical imaging (e.g., SAM-based medical models such as MedSAM2, SAM-Med3D, Med-SA) deliver appealing few‑shot behavior but struggle to enforce volumetric consistency needed for fine‑grained lesion delineation~\cite{ma2025medsam2segment3dmedical,wang2024sammed3dgeneralpurposesegmentationmodels,wu2023medicalsamadapteradapting}. Third, \textbf{language‑guided methods} (MedCLIP~\cite{wang2022medclipcontrastivelearningunpaired}, BiomedCLIP~\cite{zhang2025biomedclipmultimodalbiomedicalfoundation}) leverage biomedical vision‑language models (VLMs) to inject text priors; however, their pretraining commonly emphasizes image‑level semantics, offering organ‑level grounding without the localized, voxel‑wise guidance necessary for precise lesion boundaries. 

Recent prostate-specific studies have explored increasingly sophisticated architectures, 
including cascaded CNNs for joint prostate and dominant lesion segmentation~\cite{Eidex2022}, 
multi-stage segmentation pipelines with enhanced contextual modeling~\cite{Jacobson2025}, 
and attention-driven designs that integrate multi-scale channel and spatial cues for improved boundary sensitivity~\cite{Ding2023, Zaridis2024}. 
While these approaches demonstrate meaningful gains over earlier baselines, 
they remain fundamentally constrained by \textbf{purely visual representations} and fixed fusion strategies, 
lacking explicit semantic grounding of lesion concepts and offering limited mechanisms to adaptively regulate fusion reliability across modalities and slices. Consequently, a gap persists between robust multimodal fusion and \textbf{locally precise, lesion‑aware} guidance in 3D.

This work addresses that gap with a newly designed text-guided, multi‑encoder U‑Net that introduces lesion semantics at the bottleneck and performs late, confidence‑aware boundary edits. A bottleneck similarity head aligned to a BiomedCLIP embedding produces an interpretable heatmap used by two complementary objectives: a foreground‑only alignment loss that lifts text–image similarity inside lesions, and a probabilistic heatmap loss that suppresses spurious background activations. A confidence‑gated cross‑attention refiner is then attached only at the last decoder stage, applying localized edits where the base decoder is already reliable. Training is organized as a phase‑scheduled curriculum—segmentation stabilization → semantic transfer → gated refinement—to maintain optimization stability and to decouple semantic grounding from boundary correction. Operating exclusively at the bottleneck and last decoder layer keeps the method computationally light and easily pluggable into existing pipelines. 

Empirically, the approach yields consistent gains in overlap and surface agreement on the PI‑CAI benchmark relative to strong medical‑imaging baselines (including nnU‑Net, transformer U‑Nets, SegResNet, and SAM‑based models), and ablations confirm that both the auxiliary objectives and \textbf{phase scheduling} are essential to realizing these improvements. These results suggest a practical path to unifying multimodal fusion with localized text guidance for 3D lesion segmentation. The \textbf{contributions} are summarized next:

\begin{figure*}[t]
    \centering
    \includegraphics[width=\columnwidth, trim= 650 300 650 280]{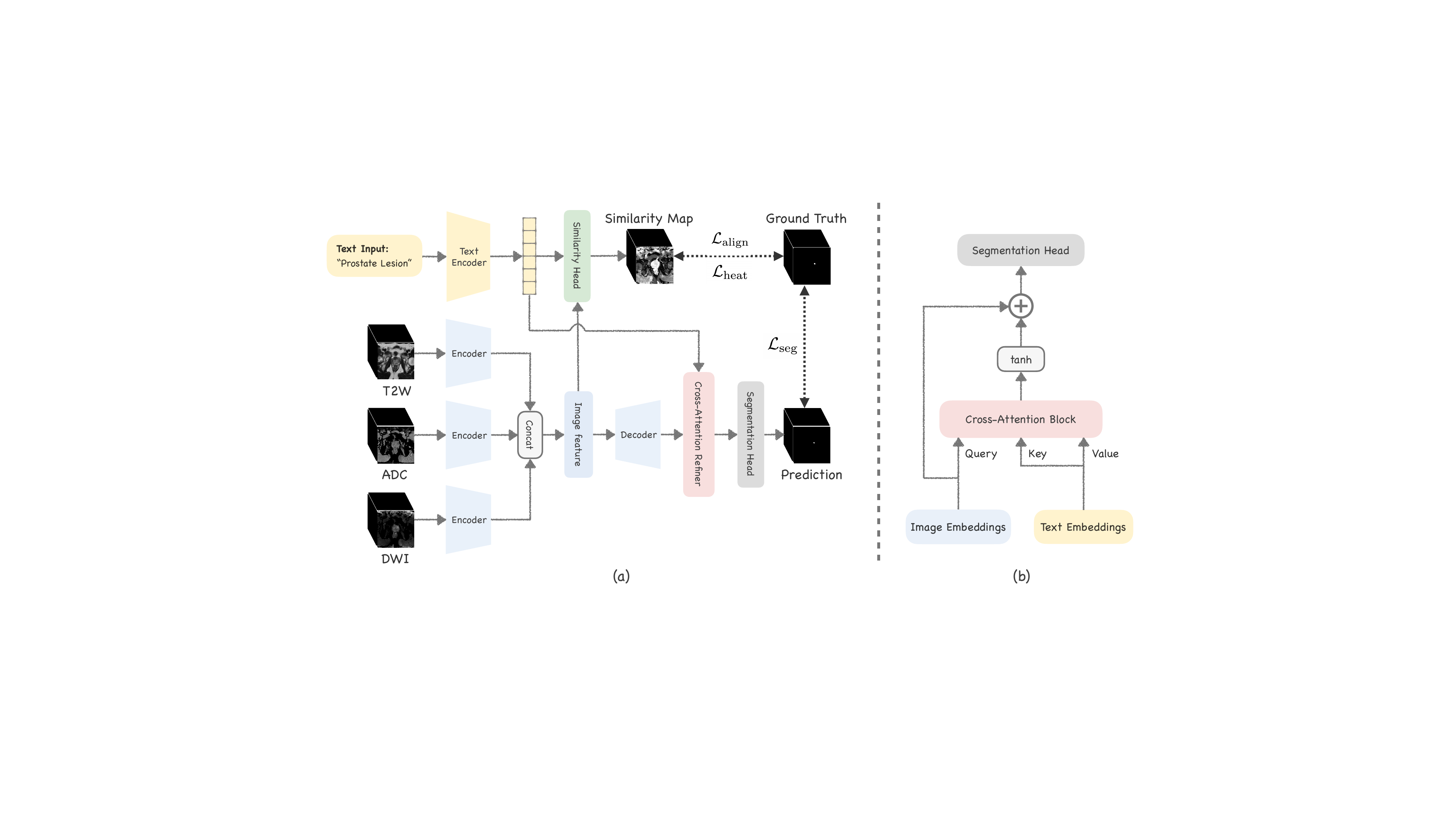}
    % \centerline{\includegraphics[height=8.7cm]{model.pdf}}
    % \vspace{-0.2cm}
    \caption{(a) Overview of proposed architecture. (b) Cross-attention refiner architecture ($Q$ from image tokens, $K/V$ from text embeddings) with a gated residual update.}
    % Multimodal inputs are encoded separately, concatenated at the highest-resolution stage, and decoded. A \emph{Similarity Head} combines text embeddings to produce an interpretable similarity map, supervised by $\mathcal{L}_{\text{align}}$ and $\mathcal{L}_{\text{heat}}$. (b) A lightweight \emph{Cross-Attention} refiner operates at the top decoder stage ($Q$ from image tokens, $K/V$ from text embeddings), applying a gated residual update before the shared segmentation head.}
    \label{fig:model}
    % \vspace{-0.2 cm}
\end{figure*}

\begin{enumerate}
\item \textbf{Framework.} Introduce a volumetric multimodal, text-guided segmentation framework for bp‑MRI, with modality-specialized encoders and a bottleneck similarity head that grounds lesion-level semantics.
\item \textbf{Bottleneck objectives.} Develop two complementary objectives at the bottleneck: (i) a foreground-only alignment loss $\mathcal{L}_{\text{align}}$ that increases text–image similarity within lesions; and (ii) a probabilistic heatmap loss $\mathcal{L}_{\text{heat}}$ that calibrates the similarity map by suppressing spurious background.
\item \textbf{Localized refinement.} Propose a confidence-gated cross-attention refiner attached only at the final decoder stage, performing small, controllable boundary edits restricted to high-confidence regions.
\item \textbf{Training curriculum.} Employ a phase-scheduled regimen—segmentation stabilization $\rightarrow$ semantic transfer $\rightarrow$ gated refinement—that improves optimization stability and decouples grounding from boundary corrections.
\item \textbf{Empirical evidence.} On PI‑CAI, our method achieves \textbf{Dice 0.7326} and \textbf{NSD 0.7541}, surpassing nnU‑Net and several transformer/SAM‑based baselines; ablations confirm the necessity of phase scheduling. 
\end{enumerate}

\section{Methods}
\textbf{Overview.} We address volumetric lesion segmentation on PI-CAI benchmark~\cite{saha2024artificial} using bp-MRI (T2W, ADC, DWI). Our multi-encoder U-Net processes each modality separately, fuses features, and decodes with a unified decoder. Text guidance from frozen BiomedCLIP with prompt “\textit{prostate lesion}” computes bottleneck similarity heatmaps and guides final-stage cross-attention refinement. Phase-scheduled training: (1) segmentation-only for multimodal fusion; (2) auxiliary losses with warm-up for lesion awareness; (3) auxiliary-disabled refinement. Architecture shown in \textbf{Fig.~\ref{fig:model}}.

\subsection{Similarity Head and Heatmap Construction}
\label{ssec:construction}
At the bottleneck, we compute a volumetric text--image similarity heatmap $s$.
Let $f\in\mathbb{R}^{B\times C\times H\times W\times D}$ (with \(B\)=batch size, \(C\)=channels, \(H\)=height, \(W\)=width, \(D\)=depth) denote the fused bottleneck features and $t\in\mathbb{R}^{B\times d}$ (with \(B\)=batch size, \(d\)=text-embedding dimension) the BiomedCLIP text embedding. We project $f$ to the text space with a $1{\times}1{\times}1$ convolution and $L_2$-normalize both representations,
$\tilde{f}=\mathrm{norm}(W_f * f),\ \tilde{t}=\mathrm{norm}(t)$. We compute a temperature‑scaled cosine similarity between normalized bottleneck features and the text embedding and apply a sigmoid to obtain an interpretable lesion‑likelihood heatmap, $s\in [0,1]$:
\[
s=\sigma\!\big(\langle \tilde{f},\tilde{t}\rangle/T\big)\in[0,1]^{H\times W\times D},
\]
where $\langle\cdot,\cdot\rangle$ denotes a per-voxel dot product with $t$ broadcast across spatial locations.
When needed, $s$ is upsampled to the segmentation resolution. The heatmap aligns image features with the text embedding and serves as a lesion-evidence map for the auxiliary losses and for training-time interpretability.

\subsection{Alignment Loss}
\label{Lalign}
We encourage high similarity inside lesions while leaving background unconstrained. We compute the loss voxel-wise at the resolution of the similarity heatmap $s$, and the ground-truth mask $M$ is resampled to match. 
We maximize the foreground average of $s$:
\[
\mathcal{L}_{\mathrm{align}}
% \mathcal{L}_{\mathrm{align}}
= 1 - \frac{\sum_{i} s_i\,M_i}{\sum_{i} M_i + \varepsilon},\
\]
where $\varepsilon$ is a small constant for numerical stability.

As a foreground-only objective, $\mathcal{L}_{\mathrm{align}}$ is computed only on lesion voxels, so it does not impose negative penalties on background. This stabilizes early optimization and better accommodates small lesions. It provides a text-to-vision semantic transfer complementary to the probabilistic heatmap loss.

\subsection{Heatmap Loss}
\label{Lheat}
To calibrate the similarity map as a lesion probability, we supervise the
similarity heatmap $s$ with the ground-truth mask $M$ using binary cross-entropy,
computed at the resolution of $s$ with $M$ resampled to match:
\[
\mathcal{L}_{\mathrm{heat}}
= -\frac{1}{N}\sum_{i=1}^{N}\big[M_i\log s_i + (1-M_i)\log(1-s_i)\big],
\]
where $N=H_s\times W_s\times D_s$.

For stability, $s$ is clamped to $[10^{-6},\,1-10^{-6}]$ before the logarithm. $\mathcal{L}_{\mathrm{heat}}$ calibrates the similarity field by suppressing bright background that reducing false positives and yields an interpretable, well-shaped alignment map that complements the foreground-only objective above.

While both auxiliary objectives operate on the same similarity map, they impose qualitatively different optimization behaviors. The alignment loss acts as a foreground-only semantic constraint, producing strictly positive gradients within lesion regions. This design stabilizes optimization under extreme class imbalance and encourages lesion-consistent feature representations. In contrast, the heatmap loss supervises the similarity map as a probabilistic lesion-likelihood map over the entire volume, introducing both positive and negative gradients to suppress spurious background activations. Together, these two objectives regularize the representation space and spatial calibration in complementary ways.

% \subsection{Text-Guided Representation Learning}
% In fine-grained prostate lesion segmentation, training with segmentation loss alone is often unstable due to extreme class imbalance and heterogeneous appearance across MRI modalities. We introduce text-guided auxiliary supervision to provide representation-level regularization. Specifically, the text embedding serves as a global semantic anchor, and the similarity head projects the shared image feature into a similarity map with respect to this semantic reference. The alignment loss $\mathcal{L}_{\mathrm{align}}$ enforces semantic consistency in the feature space, while the heatmap loss $\mathcal{L}_{\mathrm{heat}}$ imposes spatial regularization on the similarity representation. Although these auxiliary losses do not directly supervise the segmentation output, their gradients propagate through the shared encoders and reshape the feature distribution that the decoder learns from. As a result, the segmentation branch is trained on a semantically anchored and spatially regularized feature space, improving optimization stability and lesion delineation.

\subsection{Final-Stage Cross-Attention Refinement}
\label{attn}
We attach a cross-attention refiner to the highest-resolution decoder stage, leaving the backbone unchanged. Given the top-level feature map $F\in\mathbb{R}^{B\times C\times H\times W\times D}$, we flatten it into $T$ tokens and project to a hidden size $d$ to form queries $Q$. The text embedding $t
% \in\mathbb{R}^{B\times d}
$ is globally pooled and linearly projected to keys $K$ and values $V$.

The cross-attention operates on decoder features at a reduced spatial resolution, resulting in a manageable number of tokens. The feature map size is substantially smaller than the input volume due to encoder downsampling, and the text input is represented by a single global embedding. Consequently, the attention complexity remains tractable and does not incur the prohibitive memory cost associated with voxel-wise transformers.

\textbf{Refinement update.}
A cross-attention step yields a corrective update $\Delta$, which is mapped back to the original channel space and added to $F$ under a learnable scalar gate $g$ as $F' \;=\; F \;+\; g\,\Delta$, where $g=\tanh(\gamma)\in(-1,1)$ acts as a learnable gate, and $\gamma$ parameterizes the gate and thus controls the residual magnitude. The refined features pass through the segmentation head to obtain logits $y_{\text{ref}}$.

\textbf{Gated blending in confident regions.}
To avoid over-editing and preserve calibration, we blend only within high-confidence candidate regions from the base prediction. With $p=\sigma(y_{\text{base}})$ and threshold $\tau$, define the indicator mask $M_{\text{conf}}=\mathbf{1}_{\{p>\tau\}}$, and compute
\[
    y \;=\; y_{\text{base}} \;+\; \alpha\,\big(y_{\text{ref}}-y_{\text{base}}\big)\odot M_{\text{conf}}.
\]
Here, $\tau$ controls where refinements are allowed, and $\alpha$ controls how strongly they are blended.

Although both the similarity head and the cross-attention refiner are driven by the same text embedding, they serve fundamentally different roles in the proposed framework. The similarity head operates exclusively at the bottleneck during training and provides representation-level regularization by aligning multimodal image features with a global lesion semantic anchor. Its output does not directly modulate decoder features at inference time, but instead reshapes the feature distribution learned by the shared encoders through auxiliary supervision. In contrast, the cross-attention refiner is activated only at the final decoder stage and explicitly performs test-time feature editing via gated residual updates. This clear separation prevents redundancy while enabling complementary benefits.

\subsection{Phase-Scheduled Training}
\label{phases}
We adopt a three-phase curriculum to align textual and visual features while keeping optimization stable:

\begin{figure*}[t]
    \centering
    \includegraphics[width=0.75\textwidth]
    % [width=\columnwidth]
    {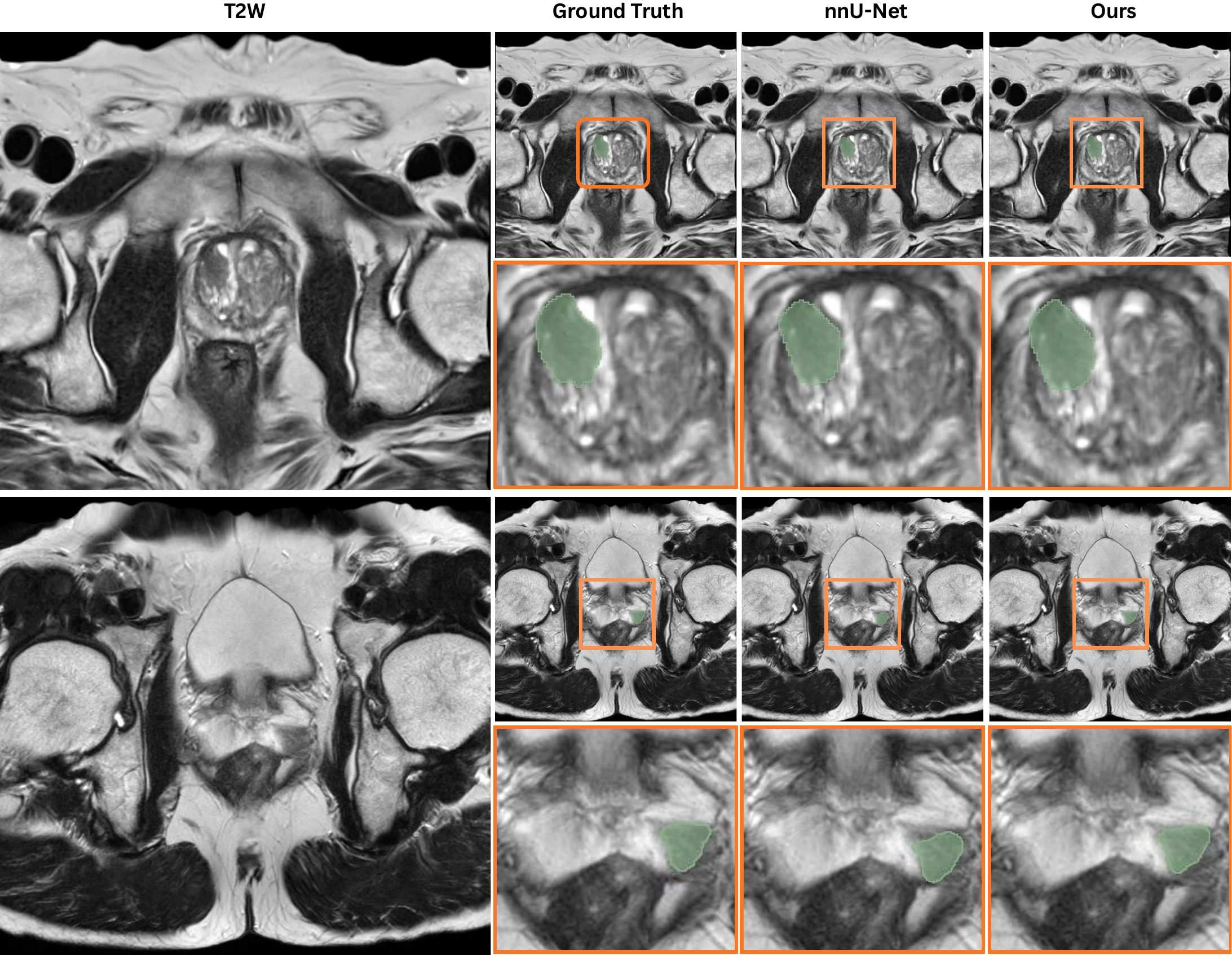}
    \caption{Qualitative comparison between nnU-Net and proposed model on bp-MRI scans.}
    % \vspace{-0.55 cm}
    \label{fig:my_pdf}
\end{figure*}

\begin{enumerate}[label=\textbf{(\arabic*)},leftmargin=*]
  \item \textbf{Segmentation-only.} Train the multi-encoder U-Net with the standard segmentation objective to stabilize multimodal fusion and establish reliable boundaries. Modality-specific encoders preserve the distinct statistical characteristics of T2W, ADC, and DWI before controlled integration at the bottleneck, thereby reducing harmful feature interference and improving feature stability under heterogeneous inputs.
  \item \textbf{Semantic transfer.} Enable two auxiliaries with a short warm-up then linear ramp: (a) the alignment objective enforces semantic consistency in the feature space by increasing text–image similarity within lesion regions; (b) the heatmap calibration objective provides spatial regularization by shaping the similarity field into a probabilistic lesion-likelihood map. Together, these objectives reshape the encoder feature distribution and strengthen lesion-specific representations.
  \item \textbf{Aux-off refinement.} Disable both auxiliaries and attach a confidence-gated, last-stage text--image cross-attention refiner that makes small local edits only where the decoder is confident, avoiding global feature perturbation while improving final delineation.
\end{enumerate}

\begin{table*}[t]
\centering
\small
\setlength{\tabcolsep}{7pt}
\caption{Comparison of SOTA methods on the PI-CAI dataset.}
\begin{tabular*}{\textwidth}{@{\extracolsep{\fill}} l c c c c c @{}}
\hline
\textbf{Model} & \textbf{Dice} & \textbf{Precision} & \textbf{Recall} & \textbf{HD95}\, & \textbf{NSD}\, \\
\hline
Attention U\mbox{-}Net~\cite{oktay2018attentionunetlearninglook} &
0.5586 $\pm$ 0.2672 & 0.6505 $\pm$ 0.3084 & 0.5654 $\pm$ 0.3131 &
29.0253 $\pm$ 38.4371 & 0.5660 $\pm$ 0.2881 \\
3D U\mbox{-}Net~\cite{3dunet} &
0.5733 $\pm$ 0.2888 & 0.6448 $\pm$ 0.3145 & 0.5811 $\pm$ 0.3203 &
30.3213 $\pm$ 36.7489 & 0.5863 $\pm$ 0.3120 \\
Swin UNETR~\cite{hatamizadeh2022swinunetrswintransformers} &
0.5811 $\pm$ 0.2951 & 0.6368 $\pm$ 0.3037 & 0.6140 $\pm$ 0.3328 &
33.2987 $\pm$ 51.7238 & 0.5858 $\pm$ 0.3103 \\
SegResNet~\cite{myronenko2018} &
0.5875 $\pm$ 0.2971 & 0.7189 $\pm$ 0.2931 & 0.5995 $\pm$ 0.3307 &
\textbf{9.6625} $\pm$ \textbf{13.2941} & 0.5910 $\pm$ 0.3248 \\
MedSAM2~\cite{ma2025medsam2segment3dmedical} &
0.5729 $\pm$ 0.1899 & 0.5071 $\pm$ 0.2176 & 0.7633 $\pm$ 0.1858 &
11.3314 $\pm$ 9.1261 & 0.7127 $\pm$ 0.1542 \\
nnU\mbox{-}Net~\cite{nnUNet} &
0.7192 $\pm$ 0.2556 & 0.7339 $\pm$ 0.2993 & \textbf{0.7864} $\pm$ \textbf{0.2249} &
24.6674 $\pm$ 47.9728 & 0.7320 $\pm$ 0.2653 \\
\hline
\textbf{Ours} &
\textbf{0.7326} $\pm$ \textbf{0.2414} & \textbf{0.7476} $\pm$ \textbf{0.2718} & 0.7795 $\pm$ 0.2655 &
15.2451 $\pm$ 28.2696 & \textbf{0.7541} $\pm$ \textbf{0.2591} \\
\hline
\end{tabular*}
\label{tab:compare}
\end{table*}

\begin{table*}[t]
\centering
\small
\setlength{\tabcolsep}{7pt}
\caption{Ablation of the proposed components on the PI-CAI dataset.}
\begin{tabular*}{\textwidth}{@{\extracolsep{\fill}} l c c c c c @{}}
\hline
\textbf{Model} & \textbf{Dice} & \textbf{Precision} & \textbf{Recall} & \textbf{HD95}\, & \textbf{NSD}\, \\
\hline
$\mathcal{L}_{\text{heat}}$ only
& 0.7131 $\pm$ 0.2644
& 0.7260 $\pm$ 0.2874
& 0.7699 $\pm$ 0.2877
& 17.6793 $\pm$ 34.5204
& 0.7252 $\pm$ 0.2830 \\

$\mathcal{L}_{\text{align}}$ only
& 0.7136 $\pm$ 0.2728
& 0.7291 $\pm$ 0.2886
& 0.7703 $\pm$ 0.2881
& 14.2987 $\pm$ 29.9708
& 0.7345 $\pm$ 0.2859 \\

$\mathcal{L}_{\text{heat}}+\mathcal{L}_{\text{align}}$
& 0.7195 $\pm$ 0.2618
& 0.7364 $\pm$ 0.2857
& 0.7645 $\pm$ 0.2904
& \textbf{13.7624} $\pm$ \textbf{29.9346}
& 0.7325 $\pm$ 0.2857 \\

$\mathcal{L}_{\text{heat}}+\mathcal{L}_{\text{align}}+\text{attn}$
& 0.7081 $\pm$ 0.2624
& 0.7214 $\pm$ 0.2879
& 0.7670 $\pm$ 0.2776
& 18.5525 $\pm$ 33.5734
& 0.7265 $\pm$ 0.2755 \\
\hline
\makecell{
$\mathcal{L}_{\text{heat}}+\mathcal{L}_{\text{align}}+\text{attn}$\\[-2pt]
\footnotesize (phase-scheduled)
}
& \textbf{0.7326} $\pm$ \textbf{0.2414}
& \textbf{0.7476} $\pm$ \textbf{0.2718}
& \textbf{0.7795} $\pm$ \textbf{0.2655}
& 15.2451 $\pm$ 28.2696
& \textbf{0.7541} $\pm$ \textbf{0.2591} \\
\hline
\end{tabular*}
\label{tab:ablation}
\end{table*}

\section{Results}

We train and evaluate our pipeline using the PI-CAI dataset \cite{saha2024artificial}. Our input consists of three modalities—T2W, ADC, and DWI volumes—each matched with a single voxel-wise lesion mask serving as the supervision label. Evaluation is conducted using patient-level cross-validation, where in each fold 70\% of cases are used for training, 10\% for validation, and 20\% are held out for testing. During training, we sample volumetric patches of size 16×320×320 (depth × height × width) with a batch size of 2. Patches are sampled with lesion-aware balancing to ensure sufficient exposure to positive regions, mitigating extreme class imbalance commonly observed in prostate lesion segmentation. The initial learning rate is set to 1e-2, and we run 1250 epochs on a single NVIDIA A100 80 GB GPU. Empirically, we set the temperature in the similarity head as 1, and the gating hyperparameters $\tau$ as 0.35 and $\alpha$ as 0.25.

We evaluate segmentation performance with Dice, precision, recall, the 95th percentile variant (HD95), and normalized surface distance (NSD). Our model achieves consistent gains in overlap and boundary quality while maintaining strong recall. The similarity head transfers lesion semantics while the confidence-gated refiner applies localized boundary edits, forming a balanced system that improves overall segmentation quality. We present baseline comparisons followed by component ablations.

\subsection{Comparison with Representative Baselines}
A comparison against representative medical segmentation models is showed in \textbf{Table~\ref{tab:compare}}. Our model obtains higher region overlap and cleaner boundaries. It achieves Dice 0.7326 and NSD 0.7541 with HD95 15.25 mm, improving over nnU-Net (Dice 0.7192, NSD 0.7320) by +0.0134 Dice and +0.0221 NSD. Recall remains competitive at 0.7795, second only to nnU-Net’s 0.7864, and precision is higher—0.7476 compared with 0.7339 for nnU-Net.

Classical U-Nets and Swin UNETR show substantially lower overlap (Dice 0.56–0.58) with larger boundary errors. SegResNet offers the lowest HD95 but weaker Dice (0.5875), reflecting its more conservative and smoother boundary predictions that avoid extreme outlier deviations but sacrifice overall overlap. MedSAM2 yields high recall but poor precision, reducing overall Dice. In comparison, our method prioritizes balanced performance—best Dice and NSD, with strong precision and competitive recall—achieving higher overlap and reliable boundary alignment. The slightly higher HD95 primarily stems from a few challenging small-lesion cases where sharper, more expressive boundaries can produce isolated long-tail deviations; nevertheless, the overall NSD remains the highest.

\subsection{Ablation of Proposed Components}
\textbf{Table~\ref{tab:ablation}} shows the effect of auxiliary losses and confidence-gated refinement. Heatmap loss alone provides a solid baseline. Replacing it with alignment loss maintains overlap but reduces extreme boundary errors (HD95: 17.68→14.30mm), effectively curbing long-tail mistakes in small lesions with ambiguous edges. Combining both losses yields complementary benefits—best outlier robustness (HD95: 13.76mm) and improved overlap. The heatmap loss calibrates text-consistent evidence while alignment suppresses severe surface errors, addressing different failure modes.

Early activation of the cross-attention refiner degrades performance (Dice 0.7081, HD95 18.55mm), suggesting its sensitivity to early-stage representation instability. When introduced too early, attention-based refinement interferes with multimodal feature learning, leading to suboptimal convergence. Phase scheduling decouples representation grounding from boundary refinement, allowing the refiner to operate on a stabilized feature space. For non-attention variants, performance saturates earlier, and additional scheduling yields marginal gains. 

The full model achieves the best overall performance across Dice, precision, recall, and NSD. Although HD95 is slightly higher than that of the non-attention variant, this behavior is expected given the nature of the metric: while the refiner improves the majority of lesion boundaries, it may also introduce rare but extreme deviations at a small number of surface points. Since HD95 measures the 95th-percentile Hausdorff distance, it is highly sensitive to such tail behavior, explaining the observed trade-off despite consistent improvements in overlap and surface consistency metrics.

\section{Discussion}
We propose a multi-encoder U-Net with phase-scheduled, text-guided training for prostate lesion segmentation on bp-MRI. 
The design sequentially stabilizes multimodal fusion, injects lesion semantics through auxiliary objectives, and applies confidence-gated cross-attention refinement to perform localized boundary edits. 
Experiments on the PI-CAI dataset demonstrate that the proposed framework achieves state-of-the-art Dice and NSD while producing interpretable similarity heatmaps aligned with lesion structures.

Our ablation results highlight two key components: bottleneck-level semantic injection and confidence-constrained refinement. 
Both contribute consistent performance gains with minimal computational overhead. 
We further observe that the phase-scheduled curriculum is essential—introducing the refiner prematurely degrades performance, confirming the importance of stabilizing representation learning before fine-grained refinement. 
Compared against strong baselines including nnU-Net, transformer-based U-Nets, SegResNet, and SAM-based models, our method achieves the best Dice and NSD with competitive HD95. 
Notably, SegResNet exhibits lower HD95 in some cases, suggesting a trade-off between overall overlap gains and extreme outlier correction, which is consistent with our design choice of localized refinement.

\section{Conclusion}
This work introduces a text-guided, phase-scheduled segmentation framework that integrates semantic alignment and confidence-aware refinement for prostate lesion segmentation. 
The proposed architecture effectively combines global lesion semantics with localized boundary corrections, yielding strong performance and improved interpretability.

Several limitations motivate future directions. 
The current evaluation follows a patient-level partition on the PI-CAI dataset, and generalizability across centers, scanners, and acquisition protocols warrants further investigation. In this work, we employ a single fixed prompt (“prostate lesion”), and thus do not explore the sensitivity of the framework to alternative prompt formulations or prompt phrasing.
Consequently, practical factors that may affect robustness—such as prompt phrasing, temperature scaling in the similarity head, and gating hyperparameters $(\tau, \alpha)$—remain to be systematically studied. 
Future extensions may incorporate boundary-aware losses to further reduce HD95, uncertainty-conditioned gating based on entropy or ensemble variance, and prompt ensembling or learnable prompts to better understand language-guidance sensitivity. 
Broader validation with multi-seed repeats, cross-site evaluation, and scanner-wise stratification will further strengthen the method’s clinical relevance and reliability.

\section*{Compliance with ethical standards}
\label{sec:ethics}
This research study was conducted retrospectively using human subject data made available in open access by (PI-CAI). Ethical approval was not required as confirmed by the license attached with the open access data.

\section*{Acknowledgment}
This work is partially supported by NIH U01-CA268808.


\begin{thebibliography}{99}
\bibitem{Gunashekar2024}
D. D. Gunashekar \textit{et al.}, ‘Comparison of data fusion strategies for automated prostate lesion detection using mpMRI correlated with whole mount histology’, \textit{Radiation Oncology}, vol. 19, 07 2024.

\bibitem{nnUNet}
F. Isensee, P. Jaeger, S. Kohl, J. Petersen, and K. Maier-Hein, ‘nnU-Net: a self-configuring method for deep learning-based biomedical image segmentation’, \textit{Nature Methods}, vol. 18, pp. 1–9, 02 2021.

\bibitem{3dunet}
Ö. Çiçek, A. Abdulkadir, S. S. Lienkamp, T. Brox, and O. Ronneberger, ‘3D U-Net: Learning Dense Volumetric Segmentation from Sparse Annotation’, arXiv:1606.06650, 2016.

\bibitem{oktay2018attentionunetlearninglook}
O. Oktay \textit{et al.}, 
‘Attention U-Net: Learning where to look for the pancreas’, 
arXiv:1804.03999, 2018.

\bibitem{hatamizadeh2022swinunetrswintransformers}
A. Hatamizadeh, V. Nath, Y. Tang, D. Yang, H. Roth, and D. Xu, ‘Swin UNETR: Swin Transformers for Semantic Segmentation of Brain Tumors in MRI Images’, arXiv:2201.01266, 2022.

\bibitem{hatamizadeh2021unetrtransformers3dmedical}
A. Hatamizadeh \textit{et al.}, ‘UNETR: Transformers for 3D Medical Image Segmentation’, arXiv:2103.10504, 2021.

\bibitem{ma2025medsam2segment3dmedical}
J. Ma \textit{et al.}, ‘MedSAM2: Segment Anything in 3D Medical Images and Videos’, arXiv:2504.03600, 2025.

\bibitem{wang2024sammed3dgeneralpurposesegmentationmodels}
H. Wang \textit{et al.}, ‘SAM-Med3D: Towards General-purpose Segmentation Models for Volumetric Medical Images’, arXiv:2310.15161, 2024.

\bibitem{wu2023medicalsamadapteradapting}
J. Wu \textit{et al}., ‘Medical SAM Adapter: Adapting Segment Anything Model for Medical Image Segmentation’, arXiv:2304.12620, 2023.

\bibitem{wang2022medclipcontrastivelearningunpaired}
Z. Wang, Z. Wu, D. Agarwal, and J. Sun, ‘MedCLIP: Contrastive Learning from Unpaired Medical Images and Text’, arXiv:2210.10163, 2022.

\bibitem{zhang2025biomedclipmultimodalbiomedicalfoundation}
S. Zhang \textit{et al.}, ‘BiomedCLIP: a multimodal biomedical foundation model pretrained from fifteen million scientific image-text pairs’, arXiv:2303.00915, 2025.

\bibitem{Eidex2022}
Z. A. Eidex \textit{et al.}, ‘MRI-based prostate and dominant lesion segmentation using cascaded scoring convolutional neural network’, \textit{Med. Phys.}, vol. 49, no. 8, pp. 5216–5224, Aug. 2022.

\bibitem{Jacobson2025}
L. E. O. Jacobson \textit{et al.}, ‘Prostate MR image segmentation using a multi-stage network approach’, \textit{Int. Urol. Nephrol}., Sep. 2025.

\bibitem{Ding2023}
M. Ding, Z. Lin, C. H. Lee, C. H. Tan, and W. Huang, 
‘A multi-scale channel attention network for prostate segmentation,’ 
\textit{IEEE Trans. Circuits Syst. II: Express Briefs}, vol. 70, no. 5, pp. 1754--1758, May 2023.


\bibitem{Zaridis2024}
D. I. Zaridis \textit{et al.}, ‘ProLesA-Net: A multi-channel 3D architecture for prostate MRI lesion segmentation with multi-scale channel and spatial attentions’, \textit{Patterns}, vol. 5, no. 7, p. 100992, Jul. 2024.



\bibitem{saha2024artificial}
A. Saha \textit{et al.}, ‘Artificial intelligence and radiologists in prostate cancer detection on MRI (PI-CAI): an international, paired, non-inferiority, confirmatory study’, \textit{Lancet Oncology}, vol. 25, no. 7, pp. 879--887, Jul. 2024.

\bibitem{myronenko2018}
A. Myronenko, ‘3D MRI brain tumor segmentation using autoencoder regularization’, arXiv:1810.11654, 2018.

\end{thebibliography}
\end{document}